\pdfoutput=1
\documentclass[11pt]{article}

\usepackage[utf8]{inputenc}
\usepackage[T1]{fontenc}
\usepackage{helvet}

\usepackage[margin=1in]{geometry}
\usepackage{amsmath,amssymb}
\usepackage{graphicx}
\usepackage{float}
\usepackage[font={small,sf},labelfont=bf]{caption}
\usepackage{algorithm}
\usepackage{algpseudocode}
\usepackage[numbers,super,sort&compress]{natbib}

\usepackage{xcolor}
\usepackage{hyperref}
\hypersetup{colorlinks=true,citecolor=blue,linkcolor=black,urlcolor=blue}

\usepackage{setspace}

\title{DoseBridge: Denoising Diffusion Bridge Model for Dose Prediction in Lung Intensity-Modulated Proton Therapy}
\author{Zerun Zhang, Xiaoda Cong, Xiangkun Xu, Peter Y. Chen and Xuanfeng Ding\thanks{%
  \textbf{Correspondence:} Xuanfeng Ding, Corewell Health William Beaumont University
  Hospital, Royal Oak, Michigan, USA.
  Email: \href{mailto:xuanfeng.ding@corewellhealth.org}{xuanfeng.ding@corewellhealth.org}}}
\date{}

\begin{document}
\singlespacing
\maketitle

% =====================================================================
\begin{abstract}
\noindent\textbf{Background.}\quad
Deep learning has enabled voxel-level radiotherapy dose prediction directly from
patient anatomy. However, most existing models condition only on the planning CT
and delineated structures, whereas the dose distribution in intensity-modulated
proton therapy (IMPT) also depends strongly on plan-specific beam geometry, which
cannot be uniquely recovered from anatomy alone. Proton dose prediction is
further constrained by the limited size of clinically curated datasets.

\noindent\textbf{Purpose.}\quad
To introduce DoseBridge, a denoising diffusion bridge model that
incorporates explicit beam geometry for accurate lung IMPT dose prediction using
a limited clinical dataset.

\noindent\textbf{Methods.}\quad
DoseBridge formulates dose prediction as a stochastic bridge between the dose
distribution and the patient CT, so that reverse generation starts from the
patient's own anatomy instead of generating the dose from Gaussian noise. Beam
geometry is encoded as a spatially aligned binary beam mask. The clinical
target volume (CTV), organs at risk, and beam mask are each represented by a
binary mask and a signed distance
map, and are fused into the denoising U-Net by a gate-fusion module at every
encoder scale and a convolutional fusion module at the bottleneck, adding only
$1.95\%$ additional parameters. CT images and treatment plans from 52 patients
with advanced-stage lung cancer (60~Gy in 30 fractions) were retrospectively
collected at a single institution; 42 cases were used for training and 10 for
testing. DoseBridge was evaluated using image-similarity metrics, clinical
dose--volume metrics, and Lyman--Kutcher--Burman normal-tissue complication
probability (NTCP) metrics and was compared with two other deep-learning
dose-prediction models.

\noindent\textbf{Results.}\quad
In the test cohort ($n=10$), DoseBridge achieved a mean absolute error of
$4.170$~Gy, a peak signal-to-noise ratio of $23.06$~dB, and a structural
similarity index of $0.798$, outperforming
both comparison models on all three metrics. The predicted CTV $D_{95}$
differed from the ground truth by
$0.62\pm1.6$~Gy. For the organs at risk, the signed differences in mean dose were
$0.14\pm1.9$~Gy (lung), $-0.32\pm2.7$~Gy (heart), $-0.03\pm1.0$~Gy (spinal cord),
and $0.24\pm2.5$~Gy (esophagus), and lung $V_{20}$ differed by $-0.08\pm3.8$
percentage points. Predicted NTCP agreed with the ground truth to within
$0.52\pm3.4$ and $-0.40\pm2.2$ percentage points for grade~2+ radiation
pneumonitis and grade~2+ acute esophagitis, respectively. Changing only the beam
mask redirected the predicted low-dose entrance regions to follow the specified
beam directions while leaving the high-dose region on the target unchanged.

\noindent\textbf{Conclusions.}\quad
DoseBridge, to our knowledge the first denoising diffusion bridge model for dose
prediction, produced IMPT dose distributions within clinically acceptable
deviations and responded to user-specified beam arrangements, supporting rapid
plan exploration under realistically limited clinical data.
\end{abstract}

\noindent\textbf{Keywords:}\quad Proton Therapy, Machine Learning, Treatment Planning

% =====================================================================
\section{Introduction}

\subsection{Background}

Radiotherapy dose prediction aims to estimate the achievable three-dimensional
dose distribution of a treatment plan directly from patient-specific anatomical
information, such as the delineated target volume and the surrounding organs at
risk (OARs). By providing a voxel-level estimate of the dose that an experienced
planner could achieve for a given anatomy, it serves as a quick, quantitative prior for
automated treatment planning, plan quality assessment, dose-guided optimization,
and adaptive radiotherapy~\cite{Wang2020review,Momin2021review}.
The resulting dose distribution can also support downstream estimation of
normal-tissue complication probability (NTCP)~\cite{Marks2010quantec}.
Compared with
conventional workflows that rely on iterative manual parameter tuning, precise
dose prediction improves planning efficiency and inter-planner consistency,
thereby supporting a smoother and more efficient clinical workflow~\cite{Fan2019,Sonke2019}.

Data-driven dose prediction was initially dominated by knowledge-based planning,
which infers one-dimensional dose--volume histogram statistics from hand-engineered
geometric features and generalizes poorly across treatment sites and
modalities~\cite{Momin2021review}. The advent of deep learning enabled end-to-end
voxel-level prediction, mapping CT images and structure masks directly to
volumetric dose; U-Net--based architectures~\cite{Ronneberger2015} became the
dominant backbone owing to their ability to resolve steep dose gradients at
target--OAR interfaces~\cite{Kearney2018,Nguyen2019hn,Babier2021openkbp}, and
conditional generative adversarial networks were subsequently introduced to
produce sharper, more realistic distributions~\cite{Mahmood2018,Babier2020gan}.
More recently, denoising diffusion probabilistic models~\cite{Ho2020,Song2021sde}
have emerged as a particularly powerful class of generative models: they offer
stable, likelihood-grounded training, surpass GANs in synthesis
quality~\cite{Dhariwal2021}, and, by sampling from the full conditional
distribution rather than regressing to its mean, preserve the sharp spatial
structure that pixel-wise losses tend to blur, motivating their rapid adoption in
medical imaging and dose prediction~\cite{Kazerouni2023survey}.

\subsection{Motivation}

Recent deep-learning methods have demonstrated the feasibility of predicting
radiotherapy dose distributions from patient anatomy. However, most existing
approaches condition their predictions primarily on the planning CT and
anatomical structures, such as the target volume and organs at
risk~\cite{Kearney2018,Nguyen2019hn,Babier2021openkbp,Zhang2024dosediff}. In
intensity-modulated proton therapy (IMPT), the dose distribution depends not only
on patient anatomy but also strongly on plan-specific beam geometry, including
the beam directions and entrance paths~\cite{Lomax1999,Cao2015beamangle}. Beam
geometry is a property of the treatment plan and cannot be uniquely recovered
from the CT or delineated anatomical structures alone. When a model is
conditioned only on anatomy, a key plan variable governing the spatial dose
distribution therefore remains unspecified, leaving the model to learn implicit
associations between patient anatomy and institution-specific beam arrangements
from the training cases. This limitation motivates the explicit encoding of beam
geometry as a spatially aligned conditioning input that can be specified using
information routinely available during treatment planning.

At the same time, proton dose prediction is constrained by the availability of
high-quality, clinically curated data. Each usable radiotherapy case requires
expert delineation of the target and organs at risk, together with a clinically
approved, dose-calculated treatment plan. Even in the more data-rich photon
setting, dose-prediction models are commonly developed using single-institution
cohorts of only a few tens to a few hundred
patients~\cite{Kearney2018,Nguyen2019hn,NguyenProstate2019,Fan2019}. This
constraint is particularly pronounced for IMPT because proton facilities remain
limited in number and treat only a small fraction of radiotherapy
patients~\cite{Zhang2026proton,Hartsell2024proton}, while proton dose prediction
lacks a public benchmark comparable to OpenKBP~\cite{Babier2021openkbp}.
Although pooling cases across institutions can increase the sample size, it also
introduces heterogeneity in contouring practices, treatment-planning styles, and
clinical conventions~\cite{Nelms2012planvariation,Vinod2016review}, while patient
privacy and regulatory requirements further restrict data
sharing~\cite{Willemink2020,Rieke2020}. When available training data are limited,
neural network models typically exhibit reduced predictive
performance~\cite{Sun2017data,Mgboh2027robustness}. Therefore, a clinically
practical proton dose-prediction method should not presume access to a large,
homogeneous multi-institutional cohort, but should instead make effective use of
task-specific structural and treatment-planning priors within the limited
clinical datasets that are realistically available.

Together, these considerations motivate the use of two complementary priors:
an anatomical prior from the patient CT and a plan-specific prior from explicit beam geometry. DoseBridge combines these priors by using
the CT as a structured bridge endpoint and incorporating a spatially aligned
beam mask through lightweight multi-scale fusion. This design aims to produce
accurate and clinically meaningful IMPT dose predictions from the realistically
limited clinical data available.

Based on these considerations, this study makes three principal contributions.
First, we propose DoseBridge, which, to our knowledge, is the first denoising
diffusion bridge model (DDBM) for radiotherapy dose prediction. DoseBridge uses the
patient CT as a structured bridge endpoint and formulates dose prediction as an
anatomically aligned CT-to-dose translation process. Second, we introduce an
explicit beam-geometry prior constructed from routine treatment-planning
parameters: the beam isocenter coordinates together with the gantry and couch
angles define the entrance path of each treatment field, and these paths are
automatically rasterized on the CT grid to form a spatially aligned binary beam
mask, which is incorporated through lightweight multi-scale fusion. Third, we
evaluate DoseBridge on a retrospective single-institution lung IMPT cohort using
voxel-level image-similarity metrics, clinical dose--volume metrics, and NTCP,
and compare its performance with
existing dose-prediction baselines.

% =====================================================================
\section{Methods}

\subsection{Denoising Diffusion Bridge Model}
\label{sec:bridge}

Diffusion models are well suited to dose prediction because their
denoising-regression objective enables stable training~\cite{Ho2020,Song2021sde},
avoiding the mode collapse and hyperparameter sensitivity common in adversarial
training~\cite{Dhariwal2021}; at the same time, their iterative refinement more
accurately characterizes the steep dose gradients around the target and organs at
risk. A standard conditional diffusion model gradually corrupts the clean dose map
$x_0$ into an isotropic Gaussian through the fixed forward process
\begin{equation}
  q\!\left(x_t \mid x_0\right)
  =\mathcal{N}\!\left(x_t;\;\sqrt{\bar\alpha_t}\,x_0,\;
   \left(1-\bar\alpha_t\right)\mathbf{I}\right),
  \label{eq:diffusion}
\end{equation}
so that the terminal state $x_T$ is pure noise. At inference, the reverse process
starts from this noise and predicts $x_0$ under a patient-specific condition
$\mathbf{c}$. The CT therefore enters only as a side condition, while generation
itself must still transport an uninformative Gaussian all the way to the dose
distribution. This detour is particularly wasteful here because a dose map is a
small-foreground signal: only a limited region receives non-negligible dose,
whereas most voxels belong to the zero-dose background. Starting from noise
therefore forces the network to model this empty region throughout denoising.
Moreover, general image generation does not ordinarily require exact preservation
of the input image structure, whereas medical image generation tasks such as image
translation must preserve precise correspondence with patient anatomy. Using the
CT both to initialize the reverse bridge and as a conditioning signal therefore
provides a stronger structural prior than injecting the CT only as a condition.

\begin{figure}[t]
  \centering
  \includegraphics[width=\linewidth]{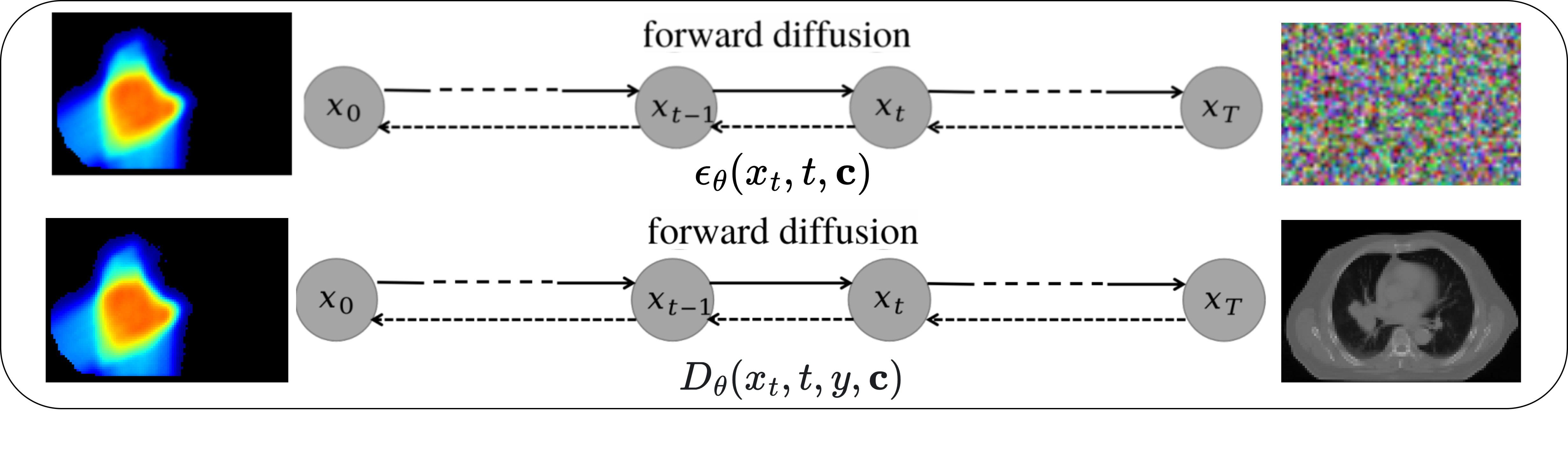}
  \caption{The \textbf{upper row} shows standard conditional diffusion, which denoises
  with $\epsilon_\theta(x_t,t,\mathbf{c})$ and has pure Gaussian noise as the terminal
  state $x_T$. The \textbf{lower row} shows the DDBM, which introduces an explicit
  bridge endpoint $y$ (the CT image), giving
  $D_\theta(x_t,t,y,\mathbf{c})$, so that the intermediate state $x_t$ lies on a
  stochastic bridge between $x_0$ and $y$ rather than on a trajectory toward pure
  Gaussian noise.}
  \label{fig:bridge}
\end{figure}

Following prior work that pins the diffusion process at both
endpoints~\cite{Li2023bbdm,Zhou2024ddbm}, the DDBM removes
this detour by replacing the Gaussian endpoint with a structured endpoint and
retaining the reference image as the starting point for denoising. Our
formulation follows the bridge construction and denoising parameterization of
the DDBM~\cite{Zhou2024ddbm}, adapted here to conditional CT-to-dose prediction.
The terminal reference image $y$ is the patient's CT, and every intermediate
state lies on a stochastic bridge between $x_0$ and $y$
(the lower row of Fig.~\ref{fig:bridge}),
\begin{equation}
  q\!\left(x_t \mid x_0,\,y\right)
  =\mathcal{N}\!\left(x_t;\;a_t\,y+b_t\,x_0,\;c_t^{2}\mathbf{I}\right),
  \label{eq:bridge}
\end{equation}
with $a_t=\bar\alpha_t\rho_t^2/\rho_T^2$, $b_t=\alpha_t\bar\rho_t^2/\rho_T^2$ and
$c_t=\alpha_t\bar\rho_t\rho_t/\rho_T$, where $\alpha_t$ and $\rho_t$ are the
signal and noise scales of a variance-preserving schedule~\cite{Song2021sde},
$\bar\alpha_t=\alpha_t/\alpha_T$ and $\bar\rho_t^2=\rho_T^2-\rho_t^2$. These
coefficients pin the process at both ends, $(a_0,b_0,c_0)=(0,1,0)$ and
$(a_T,b_T,c_T)=(1,0,0)$, so the trajectory starts exactly at the dose, ends
exactly at the CT, and is noise-free at both endpoints. Consequently, reverse
generation starts from the patient's CT rather than from noise. Following the DDBM,
the reverse direction is learned by a denoiser $D_\theta$ that recovers the clean
dose from the bridge state,
\begin{equation}
  \mathcal{L}_\theta
  =\mathbb{E}_{x_0,\,y,\,\mathbf{c},\,t,\,\epsilon}
   \left[\lambda_t\left\|
     D_\theta\!\left(x_t,t,y,\mathbf{c}\right)-x_0
   \right\|^{2}\right],
  \label{eq:loss}
\end{equation}
where $x_t=a_t\,y+b_t\,x_0+c_t\,\epsilon$ is the reparameterized bridge state of
Eq.~\eqref{eq:bridge}, with $\epsilon\sim\mathcal{N}(\mathbf{0},\mathbf{I})$, and
$\lambda_t$ is a time-dependent weighting~\cite{Zhou2024ddbm}.
Generation therefore starts from the patient's own anatomy rather than from
noise: the body contour and the exterior zero-dose background are already
provided at $t=T$ instead of being generated from scratch. This provides a more
informative starting point for the reverse generation and eases the
learning burden placed on the model.

\subsection{Beam Mask}
\label{sec:beammask}

IMPT dose distributions depend strongly on plan-specific beam geometry, which
cannot be uniquely recovered from anatomy
alone~\cite{Lomax1999,Cao2015beamangle}. We therefore expose the entrance
geometry to the network as an explicit spatial prior that captures the direction
and entrance channel of each field. This representation provides an effective
inductive bias by sparing the network from inferring the beam geometry from
the anatomical inputs alone.

To support clinical use, the beam mask is specified using routine
treatment-planning parameters. Once the beam isocenter is available from the
plan, physicians need only specify the gantry and couch angles of one or more
beams, and the corresponding mask is generated automatically. This clinically
familiar workflow supports rapid dose prediction and allows beam arrangements
to be set or perturbed when exploring candidate configurations, with the
predicted dose responding to the specified geometry. Directly feeding the raw
angles to the network, however, is
ill-suited to image-based learning: a pair of scalar angles has no spatial
correspondence with the image grid, forcing the model to infer the spatial
footprint of each beam from limited data. We therefore use an in-house script developed for the RayStation treatment planning system (RaySearch Laboratories, Stockholm, Sweden) to combine the isocenter coordinates with the gantry and couch
angles and rasterize each beam path onto the CT grid as an explicit, spatially
aligned binary mask.

Concretely, for every treatment field the script reads the isocenter coordinates
and the gantry and couch angles directly from the plan, converts the angles into
a three-dimensional beam direction, sweeps a box of fixed rectangular
cross-section along that direction, ending at the isocenter, and clips it to the
patient's body; the union of all beam boxes forms the beam mask
(Algorithm~\ref{alg:beammask}). The beam mask is rasterized onto the CT grid
and used as an additional conditioning channel.

\begin{algorithm}[h!]
\caption{Beam mask generation (RayStation)}
\label{alg:beammask}
\begin{algorithmic}[1]
\Require external body ROI $E$; beams $\{b\}$ with isocenter $p_b$ and angles
$(\gamma_b,\phi_b)$
\State $M_{\text{beam}}\gets\varnothing$
\For{each beam $b$}
    \State $d_b \gets \textsc{AngleToDir}(\gamma_b,\phi_b)$ \Comment{gantry/couch $\to$ unit vector}
    \State $B_b \gets \textsc{Box}(\text{end}=p_b,\ \text{axis}=d_b)$ \Comment{fixed rectangular cross-section}
    \State $M_{\text{beam}} \gets M_{\text{beam}} \cup (B_b \cap E)$ \Comment{clip to body}
\EndFor
\State \Return rasterize $M_{\text{beam}}$ onto the CT grid
\end{algorithmic}
\end{algorithm}

\subsection{Network Architecture}

\begin{figure}[H]
  \centering
  \includegraphics[width=\linewidth]{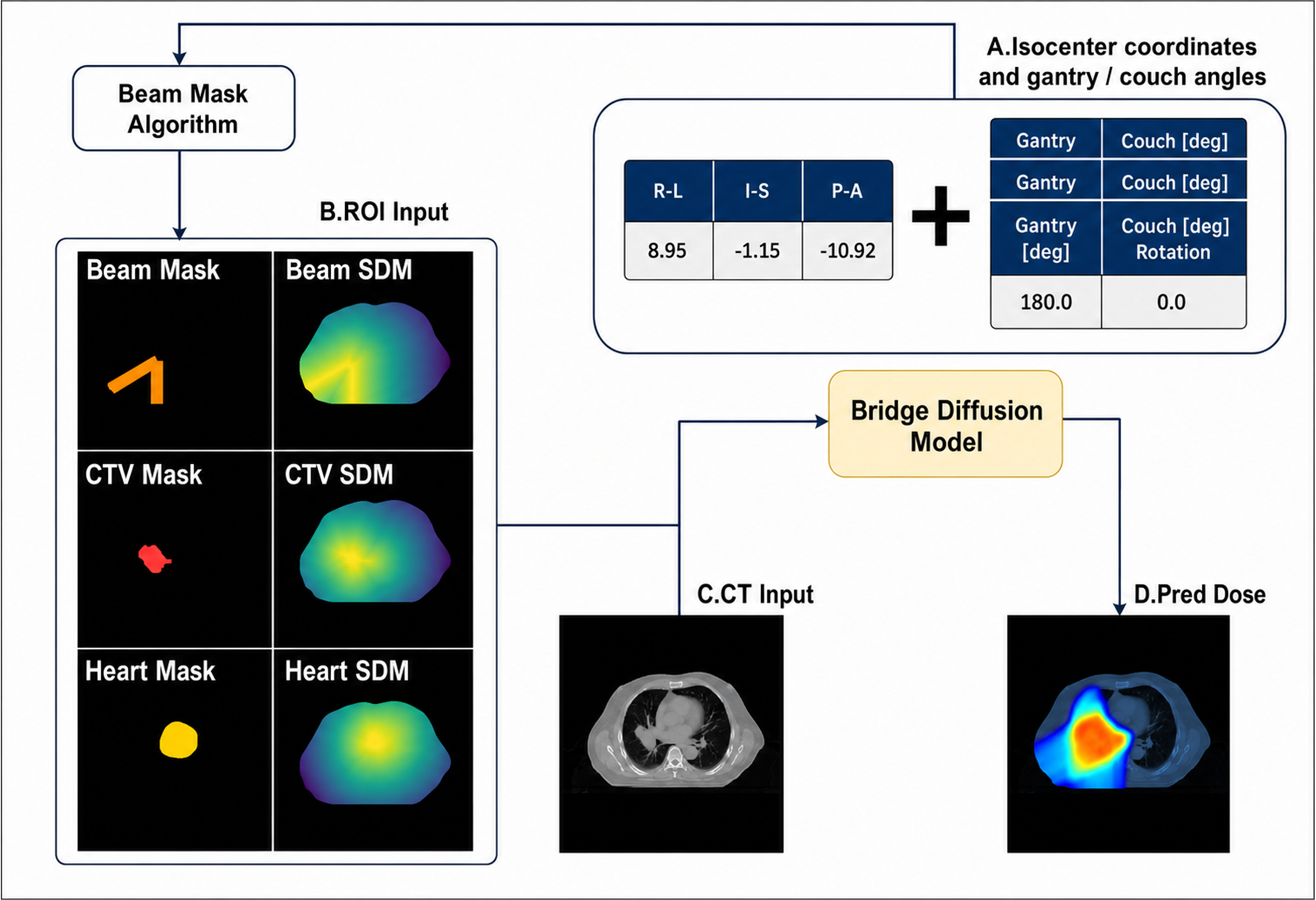}
  \caption{Conditioning inputs of the proposed model. \textbf{(A)} The beam
  isocenter coordinates, gantry angle, and couch angle are read from the
  treatment plan and \textbf{(B)} converted by the beam-mask generation
  algorithm (Algorithm~\ref{alg:beammask}) into a beam mask aligned to the CT
  space. Each conditioning structure is represented by its binary mask and its
  signed distance map, which together form the ROI stream. Only representative
  ROI structures are shown; the remaining OAR channels are omitted for clarity.
  These ROI conditions, together with \textbf{(C)} the CT image, are fed to the
  DDBM, which outputs \textbf{(D)} the predicted dose
  distribution.}
  \label{fig:inputs}
\end{figure}

\paragraph{Conditioning inputs.}
Figure~\ref{fig:inputs} summarizes the inputs to the model. Besides the
intermediate bridge state $x_t$, the network receives two conditioning streams.
The first comprises the CT image, which also serves as the bridge endpoint $y$
of Section~\ref{sec:bridge}. The second comprises the anatomical and
treatment-related structures: the clinical target volume (CTV), the OARs, and
the beam mask of Section~\ref{sec:beammask}. Following the distance-aware
conditioning of DoseDiff~\cite{Zhang2024dosediff}, each structure is represented
jointly by its binary mask and a signed distance map (SDM); the SDM assigns each
pixel its signed Euclidean distance to the corresponding structure boundary,
converting the sparse binary contours into a spatially continuous geometric
encoding. The binary masks and SDMs of all structures are stacked along the
channel dimension to form the structure stream, i.e.,\ the region-of-interest
(ROI) stream.

\begin{figure}[H]
  \centering
  \includegraphics[width=0.9\linewidth]{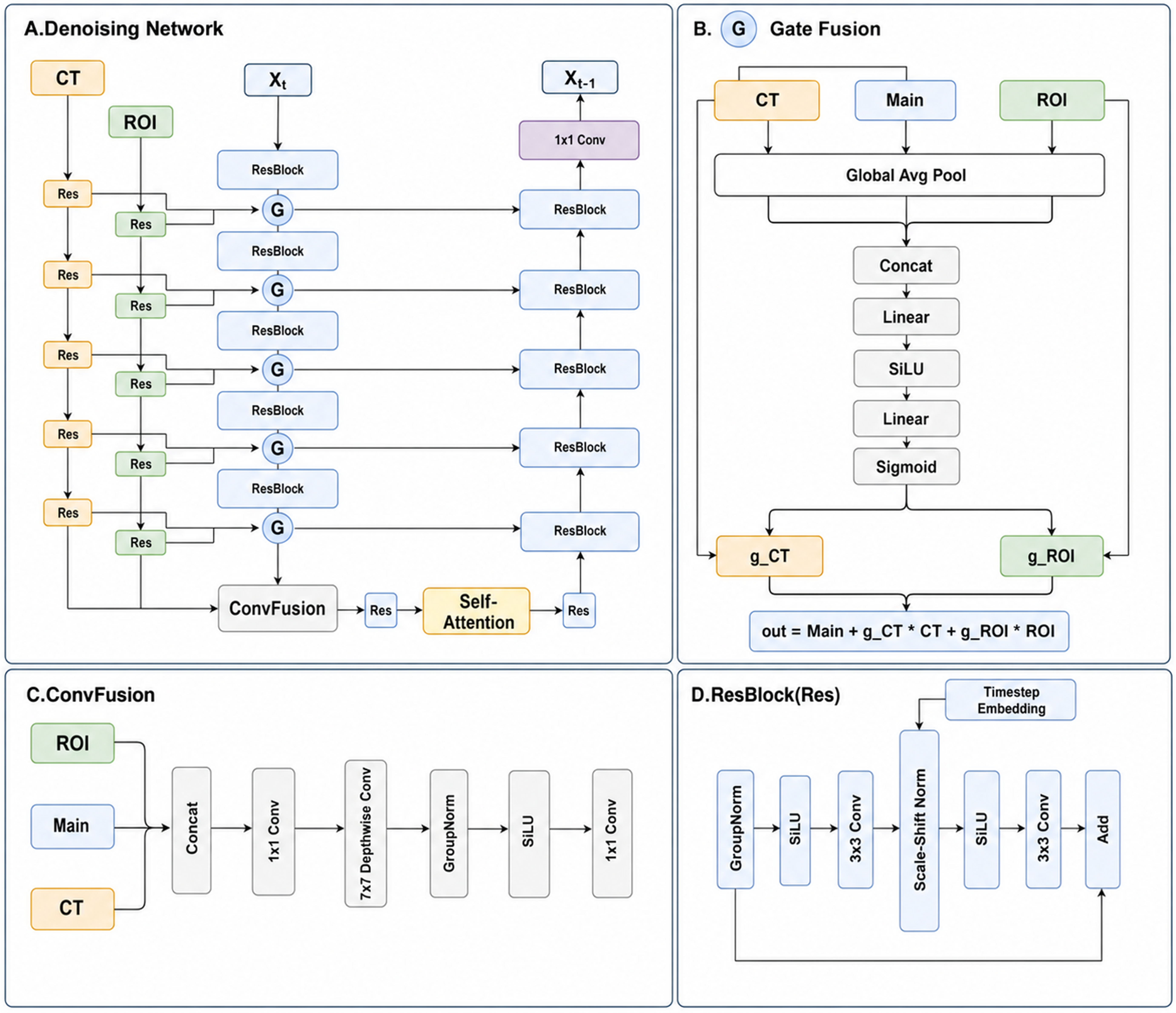}
  \caption{Architecture of the denoising network. \textbf{(A)} Three parallel
  encoders process the bridge state $x_t$, the CT stream, and the ROI stream;
  the conditions are fused into the main stream by a gate-fusion module
  (G) at every encoder scale and by a convolutional fusion module at the
  bottleneck, followed by a standard U-Net decoder. \textbf{(B)} Gate fusion:
  globally pooled descriptors of the three streams predict channel-wise sigmoid
  gates $g_{\mathrm{CT}}$ and $g_{\mathrm{ROI}}$ that weight the two conditioning
  streams before addition. \textbf{(C)} Convolutional fusion: concatenation,
  $1\times1$ reduction, $7\times7$ depthwise convolution, and a zero-initialized
  pointwise projection added residually. \textbf{(D)} Residual block with
  scale--shift timestep conditioning.}
  \label{fig:arch}
\end{figure}

\paragraph{Backbone.}
The denoiser $D_\theta$ (Fig.~\ref{fig:arch}(A)) is built upon the ADM diffusion
U-Net~\cite{Dhariwal2021}. Each stage is a residual block with scale--shift
timestep conditioning (Fig.~\ref{fig:arch}(D)). For conditional dose prediction,
the encoder comprises three parallel streams that separately process the
intermediate bridge state $x_t$, the CT image, and the ROI structures; the three
encoding streams share the same multi-scale hierarchy and
extract features through residual blocks at every resolution level. At each
encoder scale, features from the CT stream and the ROI stream are fused into the
main stream, so that anatomical and treatment-geometry information contributes to
the formation of dose features across multiple spatial resolutions. The decoder
is a standard U-Net expansion path with skip connections, and self-attention is
applied only at the coarsest feature resolutions.

\paragraph{Lightweight fusion.}
To limit the risk of overfitting under the limited clinical
dataset available here, we keep the fusion lightweight: a \emph{gate fusion}
module at every encoder scale and a \emph{convolutional fusion} module at the
network bottleneck. The gate fusion module (Fig.~\ref{fig:arch}(B)) reweights
the CT and ROI streams channel-wise in a squeeze-and-excitation
manner~\cite{Hu2018senet}: globally average-pooled descriptors of the main, CT,
and ROI streams pass through a two-layer bottleneck MLP and a sigmoid to yield
gates $g_{\mathrm{CT}}$ and $g_{\mathrm{ROI}}$, giving the fused feature
$h_{\mathrm{fused}} = h_{\mathrm{main}} + g_{\mathrm{CT}}\odot h_{\mathrm{CT}} +
g_{\mathrm{ROI}}\odot h_{\mathrm{ROI}}$. Its output weights are zero-initialized,
and the biases are set so that the gates start near one; training therefore begins from plain
additive fusion and gradually learns each condition's per-channel contribution.
At the bottleneck, the convolutional fusion module (Fig.~\ref{fig:arch}(C))
concatenates the three streams, reduces them to the original width with a
$1\times1$ convolution, mixes them spatially with a $7\times7$ depthwise
convolution in the ConvNeXt style~\cite{Liu2022convnext}, and projects the result
with a pointwise convolution. Together these modules add only about $2.67$M
parameters, or $1.95\%$ of the base network (the ADM U-Net with its CT and ROI
conditioning encoders).

% =====================================================================
\section{Experiments}

\subsection{Dataset Selection}

Fifty-two patients with advanced-stage lung cancer who were treated with IMPT
(60~Gy in 30 fractions) at our institution were retrospectively included to
evaluate the proposed DoseBridge
method. Each case provided a planning CT, associated structure delineations, a
clinically approved three-dimensional proton RBE-weighted dose distribution,
and the corresponding IMPT plan. Specifically, the clinical data for each
patient included the planning CT; delineations of the spinal cord, CTV,
esophagus, left lung, right lung, heart, and body mask; and the beam isocenter
coordinates. Ten patients were randomly held out as the test set, and the
remaining 42 were used for training.

\subsection{Experimental Setup}

The model operates in a slice-wise, two-dimensional manner: each three-dimensional
volume is processed as a stack of axial slices resampled to $128\times128$, and the
predicted slices are reassembled into the full volume. Dose values are linearly
normalized to $[-1,1]$, corresponding to $0$--$80$~Gy.
Optimization uses
AdamW (learning rate $10^{-4}$, weight decay $10^{-4}$) with gradient-norm clipping
at $1.0$ and a multi-step learning-rate decay by a factor of $0.1$ at $70\%$ of the
schedule; an exponential moving average of the weights (decay $0.9999$) is
maintained for evaluation. Training is distributed across two GPUs using PyTorch
DistributedDataParallel with a per-GPU batch size of $16$ for approximately
$10^{5}$ iterations.
At inference, dose predictions are generated efficiently using the implicit
sampler introduced by Diffusion Bridge Implicit Models
(DBIM)~\cite{Zheng2025dbim}.

\subsection{Evaluation Metrics}
\label{sec:metrics}

To assess clinical plan quality, we compare dose--volume metrics within each
region of interest. For the target we report $D_{95}$ on the original CTV mask; for the
OARs---the lung, esophagus, heart, and
spinal cord, we report structure-specific dose--volume
metrics: the mean and maximum dose ($D_{\mathrm{mean}}$ and $D_{\max}$) for the heart,
spinal cord, and esophagus, and the mean dose ($D_{\mathrm{mean}}$) together with the
volume fraction receiving at least $20$~Gy ($V_{20}$) for the lung. In addition, we use the
Lyman--Kutcher--Burman (LKB) model to assess the accuracy of the predicted
NTCP against the ground truth for two
clinically important toxicity endpoints, each modelled for its corresponding
organ at risk. For the lung, the endpoint is radiation pneumonitis of grade~2 or
higher (grade~2+), the most relevant dose-limiting lung toxicity in thoracic
radiotherapy, evaluated with the LKB parameters of Seppenwoolde et
al.~\cite{Seppenwoolde2003ntcp}. For the esophagus, the endpoint is grade~2+
acute esophagitis, the principal acute toxicity constraining esophageal dose,
evaluated with the parameters of Chapet et al.~\cite{Chapet2005esophagitis}. For
every metric, we tabulate the ground-truth value, the predicted value, and their
signed difference $\Delta=\text{Pred}-\text{GT}$, each evaluated per patient and
summarized across the test cohort. For the dose metrics, the normalized error
rate is calculated for each patient as the absolute difference between the
predicted and ground-truth values, expressed as a percentage of the 60-Gy
prescription dose; the cohort mean and standard deviation are then reported.

For comparison with prior work, we additionally use image-similarity metrics. At
the voxel level we report the mean absolute error (MAE), the peak signal-to-noise
ratio (PSNR, computed with a peak value of the $80$~Gy maximum dose), and the
structural similarity index (SSIM), the last evaluated slice-wise with a
mask-aware Gaussian window~\cite{Wang2004ssim}. Crucially, all three metrics are
computed only over voxels that lie within the body mask \emph{and} whose
ground-truth dose is at least $1$~Gy. Because a large fraction of each slice is background
or very-low-dose tissue, including these near-zero voxels would make image-based
metrics such as MAE appear artificially better, yielding overly optimistic and
poorly discriminative scores; excluding the sub-$1$~Gy region restricts the
evaluation to the clinically meaningful dose range.

% =====================================================================
\section{Results}

\subsection{Dosimetric Evaluation}

The clinical dose--volume metrics derived from the predicted and ground-truth
dose distributions over the ten test patients were summarized in
Table~\ref{tab:dvh}, while the corresponding NTCP values were summarized in
Table~\ref{tab:ntcp}. The predicted
CTV $D_{95}$ was $61.26$~Gy, compared with $60.64$~Gy for the ground truth, yielding
a signed difference ($\Delta=\text{Pred}-\text{GT}$) of $0.62\pm1.6$~Gy. After
excluding CTV-overlapping voxels from each organ at risk, the signed differences
in $D_{\mathrm{mean}}$ and $D_{\max}$ were $-0.32\pm2.7$ and $0.59\pm2.5$~Gy for
the heart, $-0.03\pm1.0$ and $1.37\pm2.3$~Gy for the spinal cord, and
$0.24\pm2.5$ and $3.74\pm6.9$~Gy for the esophagus. For the lung, the signed
differences in $V_{20}$ and $D_{\mathrm{mean}}$ were $-0.08\pm3.8$ percentage
points and $0.14\pm1.9$~Gy, respectively. Normalized to the prescription dose,
the corresponding error rates remained at or below $3.13\%$ for every
dose--volume endpoint except the esophageal $D_{\max}$ ($7.44\pm10.7\%$). The
predicted NTCP likewise tracked the ground truth closely for both toxicity
endpoints. For grade~2+ radiation pneumonitis in the lung, the signed difference
was $0.52\pm3.4$ percentage points (error rate $1.65\pm3.0\%$), and for grade~2+
acute esophagitis in the esophagus it was $-0.40\pm2.2$ percentage points (error
rate $1.26\pm1.9\%$). These results showed close agreement between the predicted
and ground-truth values for the target-coverage, organ-at-risk, and NTCP metrics
defined in Section~\ref{sec:metrics}.

\begin{table}[H]
\begin{singlespace}
\centering
\caption{Clinical dose--volume metrics of the ground-truth (GT) and predicted
(Pred) dose over the ten test patients. The rows report, from top to bottom, the cohort
mean~$\pm$~standard deviation of the GT dose, the same for the predicted dose, the
signed difference $\Delta=\text{Pred}-\text{GT}$ (cohort mean~$\pm$~standard deviation of the per-patient differences), and the
normalized error rate. Dose metrics ($D_{95}$,
$D_{\mathrm{mean}}$, $D_{\max}$) are in Gy; the volume metric $V_{20}$ is
in \%. The error rate is $|\Delta|$ normalized by the prescription dose ($60$~Gy)
for the dose metrics (Section~\ref{sec:metrics}). For $V_{20}$, the error rate is
reported as the cohort mean~$\pm$~standard deviation of the patient-level absolute
differences, in percentage points, between the predicted and ground-truth
percentages of the OAR volume (excluding the CTV) receiving at least $20$~Gy.}
\label{tab:dvh}
\footnotesize
\setlength{\tabcolsep}{4pt}
\resizebox{\textwidth}{!}{%
\begin{tabular}{l c cc cc cc cc}
\hline
 & CTV & \multicolumn{2}{c}{Heart} & \multicolumn{2}{c}{Spinal cord} & \multicolumn{2}{c}{Esophagus} & \multicolumn{2}{c}{Lung} \\
\cline{2-2}\cline{3-4}\cline{5-6}\cline{7-8}\cline{9-10}
 & $D_{95}$ (Gy) & $D_{\mathrm{mean}}$ (Gy) & $D_{\max}$ (Gy) & $D_{\mathrm{mean}}$ (Gy) & $D_{\max}$ (Gy) & $D_{\mathrm{mean}}$ (Gy) & $D_{\max}$ (Gy) & $V_{20}$ (\%) & $D_{\mathrm{mean}}$ (Gy) \\
\hline
GT       & $60.64\pm0.9$ & $5.32\pm5.0$ & $50.60\pm26.3$ & $3.00\pm3.9$ & $17.50\pm17.5$ & $8.39\pm10.4$ & $35.83\pm27.1$ & $19.02\pm10.2$ & $9.46\pm5.0$ \\
Pred     & $61.26\pm1.0$ & $5.00\pm4.8$ & $51.19\pm26.5$ & $2.97\pm3.5$ & $18.87\pm18.1$ & $8.63\pm9.3$ & $39.58\pm25.6$ & $18.93\pm11.1$ & $9.60\pm5.5$ \\
$\Delta$ & $0.62\pm1.6$ & $-0.32\pm2.7$ & $0.59\pm2.5$ & $-0.03\pm1.0$ & $1.37\pm2.3$ & $0.24\pm2.5$ & $3.74\pm6.9$ & $-0.08\pm3.8$ & $0.14\pm1.9$ \\
Err rate (\%) & $2.32\pm1.4$ & $2.58\pm3.6$ & $3.11\pm2.8$ & $1.06\pm1.1$ & $3.13\pm3.1$ & $2.71\pm3.1$ & $7.44\pm10.7$ & $2.52\pm2.7$ & $2.22\pm2.3$ \\
\hline
\end{tabular}%
}
\end{singlespace}
\end{table}

\begin{table}[H]
\begin{singlespace}
\centering
\caption{Lyman--Kutcher--Burman NTCP (\%) of the ground-truth (GT) and predicted (Pred)
dose over the ten test patients, for the two organs at risk with established LKB
parameters: grade~2+ acute esophagitis in the esophagus and grade~2+ radiation pneumonitis in the lung. The
rows report, from top to bottom, the cohort mean~$\pm$~standard deviation of the GT NTCP,
the same for the predicted NTCP, the signed difference $\Delta=\text{Pred}-\text{GT}$
(cohort mean~$\pm$~standard deviation of the per-patient differences), in percentage points, and the
NTCP error rate. The NTCP error rate is reported as the cohort
mean~$\pm$~standard deviation of the patient-level absolute differences, in
percentage points, between the LKB NTCP values computed from the predicted and
ground-truth dose distributions. The CTV is subtracted from each structure
before the gEUD is computed.}
\label{tab:ntcp}
\footnotesize
\setlength{\tabcolsep}{6pt}
\begin{tabular}{l cc}
\hline
 & Esophagus NTCP (\%) & Lung NTCP (\%) \\
\hline
GT       & $4.35\pm6.9$ & $4.32\pm4.2$ \\
Pred     & $3.95\pm5.3$ & $4.84\pm6.2$ \\
$\Delta$ & $-0.40\pm2.2$ & $0.52\pm3.4$ \\
Err rate (\%) & $1.26\pm1.9$ & $1.65\pm3.0$ \\
\hline
\end{tabular}
\end{singlespace}
\end{table}

\subsection{Method Comparison and Beam-Mask Effect}

Because no suitable proton dose-prediction model was available for direct
comparison, we selected two photon dose-prediction methods as baselines.
DoseDiff~\cite{Zhang2024dosediff} used a diffusion model conditioned on the CT
and anatomical structures, including the CTV and OARs, without explicit beam
geometry. RTPdose~\cite{Xiong2026rtpdose} was originally developed for IMRT and
VMAT dose prediction. To adapt RTPdose to the present IMPT dataset, its photon
beam mask was replaced with the proton beam mask described in
Section~\ref{sec:beammask}.

\begin{figure}[H]
  \centering
  \includegraphics[width=\linewidth]{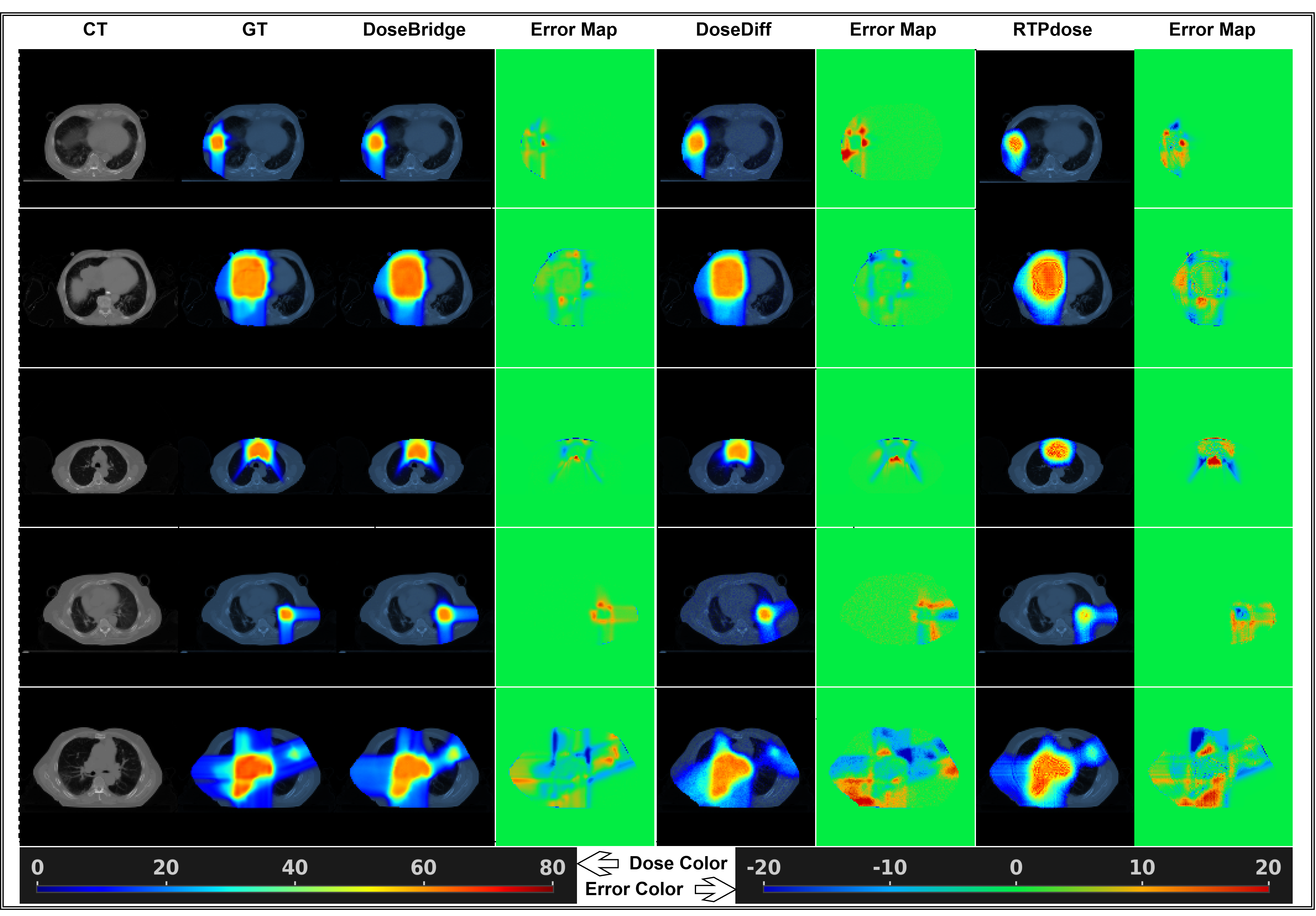}
  \caption{Qualitative comparison on five representative test slices. From left to
  right: the planning CT, the ground-truth dose, and the prediction of each method
  followed by its error map with respect to the ground truth, defined as
  $\mathrm{Error}=\mathrm{Pred}-\mathrm{GT}$. All dose maps share a common color
  scale spanning $0$--$80$~Gy, and all error maps share a common color scale
  spanning $-20$ to $+20$~Gy.}
  \label{fig:comparison}
\end{figure}

\begin{figure}[H]
  \centering
  \includegraphics[width=0.7\linewidth]{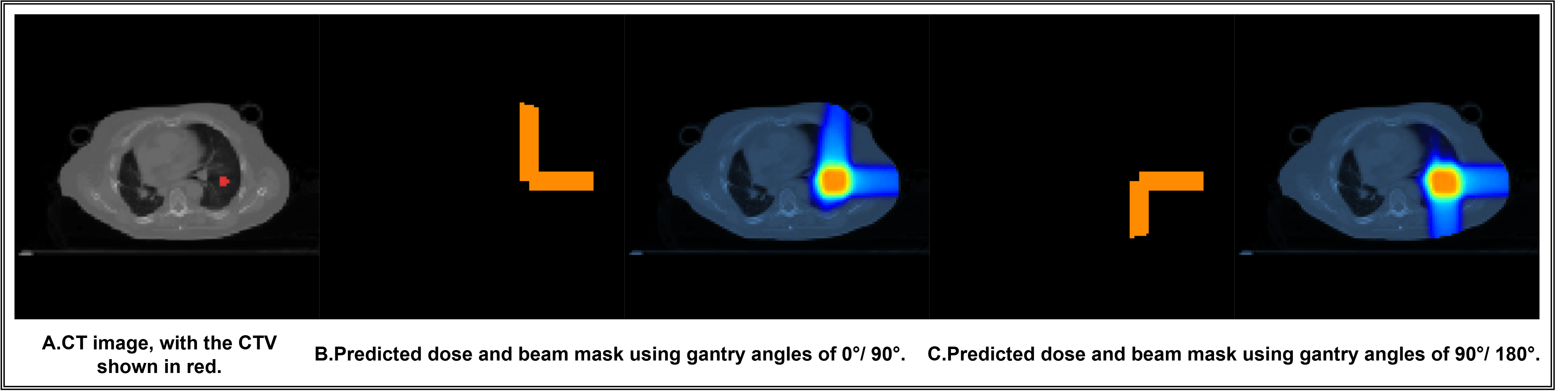}
  \caption{Qualitative effect of the beam mask on a single case. The CT and all other
  ROIs are held fixed, with only the beam mask changed, so that the differences in the
  predicted dose can be attributed solely to the beam geometry. \textbf{(A)} The
  planning CT is shown with the CTV outlined in red. \textbf{(B)} Two beams are
  used with gantry/couch angles of $(0^\circ,0^\circ)$ and $(90^\circ,0^\circ)$;
  \textbf{(C)} two beams are
  used with angles of $(90^\circ,0^\circ)$ and $(180^\circ,0^\circ)$. The low-dose regions
  along the beam paths change substantially, demonstrating the strong guiding effect
  of the beam mask on the model.}
  \label{fig:beam_ablation}
\end{figure}

As summarized in Table~\ref{tab:comparison}, DoseBridge
achieved the lowest MAE ($4.170$) and the highest PSNR ($23.06$~dB)
and SSIM ($0.798$), outperforming both photon dose-prediction baselines across
all three image-similarity metrics. This advantage was also evident qualitatively:
as shown in Fig.~\ref{fig:comparison}, DoseBridge reproduced the ground-truth dose
with the highest visual similarity among the three methods, fitting the IMPT
beam-path dose especially well. The beam-mask experiment in
Fig.~\ref{fig:beam_ablation} further showed that the beam mask exerted a strong
guiding effect on the predicted dose: the low-dose regions along the entrance
paths followed the specified beam geometry, while the high-dose region remained
on the target.

\begin{table}[H]
\begin{singlespace}
\centering
\caption{Quantitative comparison with state-of-the-art dose-prediction methods over the
ten test patients. The MAE is restricted to voxels inside the body mask
whose ground-truth dose is at least $1$~Gy (Section~\ref{sec:metrics}).}
\label{tab:comparison}
\small
\begin{tabular}{l ccc}
\hline
Method & MAE (Gy)\,$\downarrow$ & PSNR (dB)\,$\uparrow$ & SSIM\,$\uparrow$ \\
\hline
DoseDiff\cite{Zhang2024dosediff}   & $5.352$ & $21.10$ & $0.710$ \\
RTPdose\cite{Xiong2026rtpdose}     & $6.098$ & $20.38$ & $0.596$ \\
DoseBridge (ours) & $4.170$ & $23.06$ & $0.798$ \\
\hline
\end{tabular}
\end{singlespace}
\end{table}

% =====================================================================
\section{Discussion}

In this work, we proposed DoseBridge, a DDBM that
formulates IMPT dose prediction as an anatomically aligned CT-to-dose
translation, and evaluated it on a retrospective single-institution lung IMPT
cohort, where it reproduced clinical dose--volume and NTCP metrics within
clinically acceptable deviations and outperformed two state-of-the-art baselines
across all image-similarity metrics.

The performance of DoseBridge can be attributed to three design choices. First,
in contrast to standard conditional diffusion models that generate the dose from
pure Gaussian noise, DoseBridge uses the patient CT as a structured bridge
endpoint: the body contour and the exterior zero-dose background are already
provided at the start of reverse generation, so the model capacity is
concentrated on the clinically relevant dose region --- a property that is
particularly valuable under the limited clinical data available for proton
therapy. Second, the beam mask exposes plan-specific beam geometry, which cannot
be recovered from anatomy alone, as a spatially aligned prior; as shown in
Fig.~\ref{fig:beam_ablation}, the predicted low-dose entrance regions follow the
specified beam directions while target coverage is preserved, which also allows
physicians to explore candidate beam arrangements by simply specifying the
isocenter and the gantry and couch angles. Third, the gate and convolutional fusion modules inject the CT and ROI
conditions at multiple scales while adding only about $1.95\%$ additional
parameters, limiting the risk of overfitting on a small cohort.

Several limitations of this study should be acknowledged. First, the model was
developed and evaluated on a single-institution cohort of 52 lung IMPT patients
treated with a uniform prescription; its generalizability to other treatment
sites, prescriptions, and institutions with different planning styles remains to
be validated on larger multi-institutional data. Second, the esophageal maximum
dose was overestimated ($\Delta D_{\max}=3.74\pm6.9$~Gy), the only endpoint whose
normalized error rate exceeded $3.13\%$.
Because the esophagus often abuts the target, its $D_{\max}$ is determined
by a few voxels on a steep dose gradient and is therefore highly sensitive to
small spatial deviations in the prediction; notably, the esophageal
$D_{\mathrm{mean}}$ and NTCP deviations remained small ($0.24$~Gy and $-0.40$
percentage points), indicating that the overall dose burden was still captured
accurately. Third, a similar though milder behavior was observed for the spinal
cord ($\Delta D_{\max}=1.37\pm2.3$~Gy). For serial organs whose clinical
constraints are defined by point doses, the predicted dose should therefore be
regarded as a planning prior rather than a substitute for dose calculation, and
final plan approval should still rely on the treatment-planning system.

% =====================================================================
\section{Conclusions}

We proposed DoseBridge, which, to our knowledge, is the first DDBM for
radiotherapy dose prediction and is designed specifically for IMPT. DoseBridge
uses the patient CT as a structured bridge endpoint and incorporates
plan-specific beam geometry through a spatially aligned beam mask with
lightweight multi-scale fusion. On a
retrospective lung IMPT cohort, DoseBridge outperformed two state-of-the-art
baselines in image-similarity metrics and reproduced clinical
dose--volume and NTCP metrics within clinically acceptable deviations, while allowing the
predicted dose to respond to user-specified beam arrangements. These results
suggest that the DDBM combined with treatment-planning priors is a
practical route to accurate and clinically meaningful IMPT dose prediction
under realistically limited clinical data.

% =====================================================================
\section*{Declarations}

\noindent\textbf{Data availability statement.}\quad The data used in this study are restricted to internal research purposes and are not publicly available. Requests for data access may be directed to the corresponding author.

\noindent\textbf{Use of AI.}\quad The authors used artificial-intelligence tools to
assist with the revision of the manuscript and the preparation of figures.

\noindent\textbf{Acknowledgments}\quad The study is supported, in part, by NIH grant R01CA301448, IBA and Elekta research funding, and the Dr. Peter Y. Chen \& Family Fund at the Corewell Health Foundation Southeast Michigan.

\noindent\textbf{Conflict of interest statement}\quad 
Xuanfeng Ding received honoraria from IBA and the Elekta Speaker Bureau outside the scope of the work presented here. Xuanfeng Ding is an inventor on a patent related to Particle Arc Therapy, which is assigned to Corewell Health. A related license agreement exists with IBA.
% --------------------------- bibliography ----------------------------
\bibliographystyle{unsrtnat}
\bibliography{refs}

\end{document}